\documentclass[11pt,a4paper]{article}
\usepackage[hyperref]{emnlp2020}
\usepackage{times}
\usepackage{amsmath}
\usepackage{bm}
\usepackage{multirow}
\usepackage{latexsym}
\usepackage{subfigure}
\usepackage{graphicx}
\usepackage{amsfonts}
\usepackage{enumitem}
\usepackage{balance}
\usepackage{booktabs} % For formal tables

\newtheorem{definition}{Definition}[section]

\newcommand{\squishlist}{
 \begin{list}{$\bullet$}
  { \setlength{\itemsep}{0pt}
     \setlength{\parsep}{3pt}
     \setlength{\topsep}{3pt}
     \setlength{\partopsep}{0pt}
     \setlength{\leftmargin}{1.5em}
     \setlength{\labelwidth}{1em}
     \setlength{\labelsep}{0.5em} } }

\newcommand{\squishlisttwo}{
 \begin{list}{$\bullet$}
  { \setlength{\itemsep}{0pt}
     \setlength{\parsep}{0pt}
    \setlength{\topsep}{0pt}
    \setlength{\partopsep}{0pt}
\setlength{\leftmargin}{2em}
\setlength{\labelwidth}{1.5em}
\setlength{\labelsep}{0.5em} } }

\newcommand{\squishend}{
\end{list}  }

\usepackage{microtype}

\aclfinalcopy % Uncomment this line for the final submission
\def\aclpaperid{151} %  Enter the acl Paper ID here

\title{Be More with Less: Hypergraph Attention Networks for \\ Inductive Text Classification}

\author{Kaize Ding \\
  Arizona State University  \\
  \texttt{kaize.ding@asu.edu} \\\And
  Jianling Wang \\
  Texas A\&M University  \\
  \texttt{jlwang@tamu.edu} \\\And
  Jundong Li \\
  University of Virginia  \\
  \texttt{jundong@virginia.edu} \\\AND
  Dingcheng Li \\
  Amazon Inc.  \\
  \texttt{lidingch@amazon.com} \\\And
  Huan Liu \\
  Arizona State University  \\
  \texttt{huan.liu@asu.edu} \\}

\date{}

\begin{document}
\maketitle
\begin{abstract}
Text classification is a critical research topic with broad applications in natural language processing. Recently, graph neural networks (GNNs) have received increasing attention in the research community and demonstrated their promising results on this canonical task. Despite the success, their performance could be largely jeopardized in practice since they are: (1) unable to capture high-order interaction between words; (2) inefficient to handle large datasets and new documents. To address those issues, in this paper, we propose a principled model -- hypergraph attention networks (HyperGAT), which can obtain more expressive power with less computational consumption for text representation learning. Extensive experiments on various benchmark datasets demonstrate the efficacy of the proposed approach on the text classification task.
\end{abstract}

\section{Introduction}

Text classification, as one of the most fundamental tasks in the field of natural language processing, has received continuous endeavors from researchers due to its wide spectrum of applications, including sentiment analysis~\cite{wang2016attention}, topic labeling~\cite{wang2012baselines}, and disease diagnosis~\cite{miotto2016deep}. Inspired by the success of deep learning techniques, methods based on representation learning such as convolutional neural networks (CNNs)~\cite{kim2014convolutional} and recurrent neural networks (RNNs)~\cite{liu2016recurrent} have been extensively explored in the past few years. In essence, the groundbreaking achievements of those methods can be attributed to their strong capability of capturing sequential context information from local consecutive word sequences.

% Recently, deep learning
% models have been widely used to learn text representations.

% Although neural approaches for
% text classification have been quite effective

% By considering words and documents as nodes across the entire corpus, we can 

More recently, graph neural networks (GNNs)~\cite{kipf2016semi, velivckovic2017graph, hamilton2017inductive} have drawn much attention and demonstrated their superior performance in the task of text classification~\cite{yao2019graph,wu2019simplifying,liu2020tensor}. This line of work leverages the knowledge from both training and test documents to construct a corpus-level graph with global word co-occurrence and document-word relations, and consider text classification as a semi-supervised node classification problem. Then with GNNs, long-distance interactions between words could be captured to improve the final text classification performance.

% Specifically, based on the word co-occurrence and document-word relations of both training and test documents, those methods construct a single, fixed text graph across the entire corpus. 

% In this way, the text classification problem can be directly turn to a semi-supervised node classification problem.

% construct a single large graph that encodes sequential contextual relationship between words. In this way, those GNN-based models are able to capture long-distance dependency between words. Notably, those methods turn the text classification problem to a semi-supervised node classification problem, which are inherently \textit{transductive} since the constructed text graph includes both the

Despite their promising early results, the usability of existing efforts could be largely jeopardized in real-world scenarios, mainly owing to their limitations in the following two aspects: \textbf{(i)} \textit{Expressive Power.} Existing GNN-based methods predominately focus on pairwise interactions (i.e., dyadic relations) between words. However, word interactions are not necessarily dyadic in natural language, but rather could be triadic, tetradic, or of a higher-order. 
For instance, consider the idiom ``\texttt{eat humble pie}'', whose definition is ``admit that one was wrong'' in common usage. If we adopt a simple graph to model the word interactions, GNNs may misinterpret the word \texttt{pie} as ``a baked dish" based on its pairwise connections to other two words  (\texttt{humble} -- \texttt{pie} and \texttt{eat} -- \texttt{pie}), then further misunderstand the actual meaning of the whole idiom. Hence, how to go beyond pairwise relations and further capture the high-order word interactions is vital for high-quality text representation learning, but still remains to be explored. \textbf{(ii)} \textit{Computational Consumption.} On the one hand, most of the endeavors with GNN backbone tend to be memory-inefficient when the scale of data increases, due to the fact that constructing and learning on a global document-word graph consumes immense memory~\cite{huang2019text}. On the other hand, the mandatory access to test documents 
during training renders those methods inherently \textit{transductive}. It means that when new data arrives, we have to retrain the model from scratch for handling newly added documents. Therefore, it is necessary to design a computationally efficient approach for solving graph-based text classification.

%  as example: 
% In fact, we can hardly map the representation of ``\texttt{pie}'' to its real meaning ``admit that one was wrong'' by using simple graph, since ``\texttt{pie}'' in both phrases ``\texttt{humble pie}'' and ``\texttt{eat pie}'' means ``a baked dish".

% Those issues necessitate the new design of a document-level model for \textit{inductive} text classification; 

% Additionally, the importance of each word varies a lot for predicting the label of document, while the existing GNN-based methods rarely consider that. It is indispensable to highlight more informative words to achieve more accurate classification results. 

% require the prior knowledge of test data.

Upon the discussions above, one critical research question to ask is ``\emph{Is it feasible to acquire more expressive power with less computational consumption?}''. To achieve this goal, we propose to adopt document-level hypergraph (hypergraph is a generalization of simple graph, in which a hyperedge can connect \textit{arbitrary number of nodes}) for modeling each text document. The use of document-level hypergraphs potentially enables a learning model not only to alleviate the computational inefficiency issue, but more remarkably, to capture heterogeneous (e.g., sequential and semantic) high-order contextual information of each word. Therefore, more expressive power could be obtained with less computational consumption during the text representation learning process. As conventional GNN models are infeasible to be used on hypergraphs, to bridge this gap, we propose a new model named HyperGAT, which is able to capture the encoded high-order word interactions within each hypergraph. In the meantime, its internal dual attention mechanism highlights key contextual information for learning highly expressive text representations. To summarize, our contributions are in three-fold:

\begin{itemize}[leftmargin=*,noitemsep,topsep=1.5pt]% 
\item We propose to model text documents with document-level hypergraphs, which improves the model expressive power and reduces computational consumption. 
\smallskip
\item A principled model HyperGAT based on a dual attention mechanism is proposed to support representation learning on text hypergraphs.

\smallskip
\item We conduct extensive experiments on multiple benchmark datasets to illustrate the superiority of HyperGAT over other state-of-the-art methods on the text classification task.
\end{itemize}

% It is noted that the data structure in real practice could be beyond pairwise connections and even far more complicated.

\section{Related Work}
\subsection{Graph Neural Networks}
Graph neural networks (GNNs) -- a family of neural models for learning latent node representations in a graph, have achieved remarkable success in different graph learning tasks~\cite{defferrard2016convolutional,kipf2016semi,velivckovic2017graph,ding2019deep,ding2020graph}. Most of the prevailing GNN models follow the paradigm of neighborhood aggregation, aiming to learn latent node representations via message passing among local neighbors in the graph. With deep roots in graph spectral theory, the learning process of graph convolutional networks (GCNs)~\cite{kipf2016semi} can be considered as a mean-pooling neighborhood aggregation. Later on, GraphSAGE~\cite{hamilton2017inductive} was developed to concatenate the node's feature with mean/max/LSTM pooled neighborhood information, which enables inductive representation learning on large graphs. Graph attention networks (GATs)~\citep{velivckovic2017graph} incorporate trainable attention weights to specify fine-grained weights on neighbors when aggregating neighborhood information of a node. Recent research further extend GNN models to consider global graph information~\cite{battaglia2018relational} and edge information~\cite{gilmer2017neural} during aggregation. More recently, hypergraph neural networks~\cite{feng2019hypergraph,bai2020hypergraph,wang2020next} are proposed to capture high-order dependency between nodes. Our model HyperGAT is the first attempt to shift the power of hypergraph to the canonical text classification task.

\subsection{Deep Text Classification}
Grounded on the fast development of deep learning techniques, various neural models that automatically represent texts as embeddings have been developed for text classification. Two representative deep neural models, CNNs~\citep{kim2014convolutional,zhang2015character} and RNNs~\citep{tai2015improved,liu2016recurrent} have shown their superior power in the text classification task. To further improve the model expressiveness, a series of attentional models have been developed, including hierarchical attention networks~\citep{yang2016hierarchical}, attention over attention~\citep{cui2016attention}, etc. More recently, graph neural networks have shown to be a powerful tool for solving the problem of text classification by considering the long-distance dependency between words. Specifically, TextGCN~\citep{yao2019graph} applies the graph convolutional networks (GCNs)~\citep{kipf2016semi} on a single large graph built from the whole corpus, which achieves state-of-the-art performance on text classification. Later on, SGC~\citep{wu2019simplifying} is proposed to reduce the unnecessary complexity and redundant computation of GCNs, and shows competitive results with superior time efficiency. TensorGCN~\citep{liu2020tensor} proposes a text graph tensor to learn word and document embeddings by incorporating more context information. \cite{huang2019text} propose to learn text representations on document-level graphs. However, those transductive methods are computationally inefficient and cannot capture the high-order interactions between words for improving model expressive power.

\section{Methodology}
In this section, we introduce a new family of GNN models developed for inductive text classification. By reviewing the existing GNN-based endeavors, we first summarize their main limitations that need to be addressed. Then we illustrate how we use hypergraphs to model text documents for achieving the goals. Finally, we propose the model HyperGAT based on a new dual attention mechanism and model training for inductive text classification.

\subsection{GNNs for Text Classification}
With the booming development of deep learning techniques, graph neural networks (GNNs) have achieved great success in representation learning on graph-structured data~\cite{zhou2018graph,ding2019feature}. In general, most of the prevailing GNN models follow the neighborhood aggregation strategy, and a GNN layer can be defined as:
\begin{equation}
    \mathbf{h}_i^l = \textsc{Aggr}^l\Big( \mathbf{h}_i^{l-1},  \{   \mathbf{h}_j^{l-1} \arrowvert \forall j \in \mathcal{N}_i \}\Big),
    \label{eq:graphSage}
\end{equation}
where $\mathbf{h}_i^{l}$ is the node representation of node $i$ at layer $l$ (we use $\mathbf{x}_i$ as $\mathbf{h}_i^0$) and
$\mathcal{N}_i$ is the local neighbor set of node $i$. \textsc{Aggr} is the aggregation function of GNNs and has a series of possible implementations~\cite{kipf2016semi,hamilton2017inductive,velivckovic2017graph}.

Given the capability of capturing long-distance interactions between entities, GNNs also have demonstrated promising performance on text classification~\cite{yao2019graph,wu2019domain,liu2020tensor}. The prevailing approach is to build a corpus-level document-word graph and try to classify documents through semi-supervised node classification. Despite their success, most of the existing efforts suffer from the computational inefficiency issue, not only because of the mandatory access of test documents, but also the construction of corpus-level document-word graphs. In the meantime, those methods are largely limited by the expressibility of using simple graphs to model word interactions. Therefore, how to improve model expressive power with less computational consumption is a challenging and imperative task to solve.

% those transductive methods not only , but also are limited by the model expressiveness. 

% However, the mandatory access of test documents during training renders those models inherently \textit{transductive}. Also, the construction of the global corpus-level graph largely magnifies the difficulty to use them on large-scale datasets. Therefore, the \textit{computational efficiency} issue is imperative to solve.

% most of them are not computationally efficient due to two main reasons: (1)

% Therefore, those  are unable to handle new documents without retraining the model. 

% which have critical drawback in terms of \textit{Computational Cost}.

% still have critical limitations in terms of \textit{Computational Cost} and \textit{Expressive Power}. 

% Another challenge is that existing GNN endeavors use pairwise relations to model the interaction between words. How to capture the high-order word interactions encoded in the text documents. The main reason is that the learning of TextGCN is based on a fixed corpus-level graph that only models pairwise relations between words.  

\subsection{Documents as Text Hypergraphs}
To address the aforementioned challenges, in this study, we alternatively propose to model text documents with document-level hypergraphs. Formally, hypergraphs can be defined as follows:

% This way enables us to bridge the gap between \textit{Computaional Efficiency} and \textit{Expressive Power}. 

% Note that by using hypegraph, we are able to capture heterogeneous (e.g., semantic, syntactic) context information for text representation learning. 

 \begin{figure*}[t]
    \graphicspath{{figures/}}
    \centering
    \includegraphics[width=0.95\textwidth]{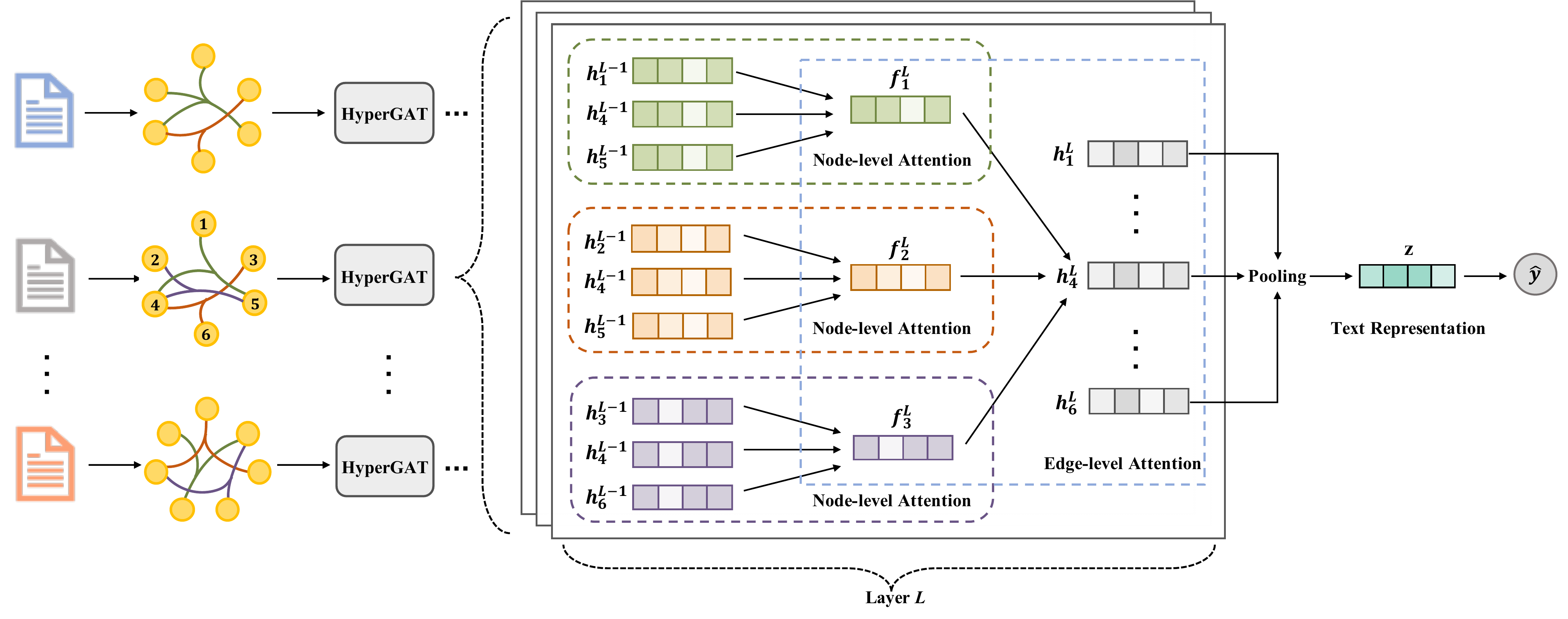}
    \caption{Illustration of the proposed hypergraph attention networks (HyperGAT) for inductive text classification. We construct a hypergraph for each text document and feed it into HyperGAT. Based on the node and edge-level attention, text representations that capture high-order word interactions can be derived. Figure best viewed in color. }%
    \label{fig:framework}%
\end{figure*}

\begin{definition}
\textbf{Hypergraphs}: A hypergraph is defined as a graph $ G = (\mathcal{V}, \mathcal{E})$, where $\mathcal{V} = \{v_1, \dots, v_n\}$ represents the set of nodes in the graph, and $\mathcal{E} = \{e_1, \dots, e_m \}$ represents the set of hyperedges. Note that for any hyperedge $e$, it can connect two or more nodes (i.e., $\sigma(e) \geq 2$).
\end{definition}

Notably, the topological structure of a hypergraph $G$ can also be represented
by an incidence matrix $\mathbf{A} \in  \mathbb{R}^{n \times m}$, with entries defined as:
\begin{equation}
    \mathbf{A}_{ij} =  \left\{ \begin{aligned}
        1, &\text{ if } v_i \in e_j, \\
        0, &\text{ if } v_i \not\in e_j.
    \end{aligned} \right.
\end{equation}

In the general case, each node in hypergraphs could come with a $d$-dimensional attribute vector. Therefore, all the node attributes can be denoted as $\mathbf{X} = [\mathbf{x}_1, \mathbf{x}_2, \dots, \mathbf{x}_n]^{\mathrm{T}} \in \mathbb{R}^{n \times d}$, and we can further use $G = (\mathbf{A}, \mathbf{X})$ to represent the whole hypergraph for simplicity. 

For a text hypergraph, nodes represent words in the document and node attributes could be either one-hot vector or the pre-trained word embeddings (e.g., word2vec, GloVe). In order to model heterogeneous high-order context information within each document, we include multi-relational hyperedges as follows:

\smallskip
\paragraph{Sequential Hyperedges.}
Sequential context depicts the language property of local co-occurrence between words, which has demonstrated its effectiveness for text representation learning~\cite{yao2019graph}. To leverage the sequential context information of each word, we first construct sequential hyperedges for each document in the corpus. One natural way is to adopt a fixed-size sliding window to obtain global word co-occurrence as the sequential context. Inspired by the success of hierarchical attention networks~\cite{yang2016hierarchical}, here we consider each sentence as a hyperedge and it connects all the words in this sentence. As another benefit, using sentences as sequential hyperedges enables our model to capture the document structural information at the same time. 

\smallskip
\paragraph{Semantic Hyperedges.}
Furthermore, in order to enrich the semantic context for each word, we build semantic hyperedges to capture topic-related high-order correlations between words~\cite{linmei2019heterogeneous}. Specifically, we first mine the latent topics $T$ from the text documents using LDA~\cite{blei2003latent} and each topic $t_i = (\bm\theta_1, ..., \bm\theta_w)$ ($w$ denotes the vocabulary size) can be represented by a probability distribution over the words. Then for each topic, we consider it as a semantic hyperedge that connects the top $K$ words with the largest probabilities in the document. With those topic-related hyperedges, we are able to enrich the high-order semantic context of words in each document.

It is worth mentioning that though we only discuss sequential and semantic hyperedges in this study, other meaningful hyperedges (e.g., syntactic-related) could also be integrated into the proposed model for further improving the model expressiveness and we leave this for future work.

\subsection{Hypergraph Attention Networks} 
To support text representation learning on the constructed text hypergraphs, we then propose a new model called HyperGAT (as shown in Figure 1) in this section. Apart from conventional GNN models, HyperGAT learns node representations with two different aggregation functions, allowing to capture heterogeneous high-order context information of words on text hypergraphs. In general, a HyperGAT layer can be defined as:
\begin{equation}
\begin{aligned}
    \mathbf{h}_{i}^l &= \textsc{Aggr}^l_{edge}\Big(\mathbf{h}_i^{l-1},\{   \mathbf{f}_j^{l} \arrowvert \forall e_j \in \mathcal{E}_i \}\Big),\\
    \mathbf{f}_j^{l} &= \textsc{Aggr}^l_{node}\Big(\{   \mathbf{h}_{k}^{l-1} \arrowvert \forall v_k \in e_j \}\Big)
    \label{eq:graphSage2},
\end{aligned}
\end{equation}
where $\mathcal{E}_i$ denotes the set of hyperedges connected to node $v_i$ and $\mathbf{f}_j^l$ is the representation of hyperedge $e_j$ in layer $l$. $\textsc{Aggr}_{edge}$ is an aggregation function that aggregates features of hyperedges to nodes and  $\textsc{Aggr}_{node}$ is another aggregation function that aggregates features of nodes to hyperedges. In this work, we propose to implement those two functions based on a dual attention mechanism. We will start by describing a single layer $l$ for building arbitrary HyperGAT architectures as follows:

% On the one hand, HyperGAT is capable of modeling high-order word interactions from the text hypergraphs; On the other hand, by virtue of a dual attention mechanism, HyperGAT highlights informative context information when learning word representations. 

\smallskip
\paragraph{Node-level Attention.} Given a specific node $v_i$, our HyperGAT layer first learns the representations of all its connected hyperedges $\mathcal{E}_i$. As not all the nodes in a hyperedge $e_j \in \mathcal{E}_i$ contribute equally to the hyperedge meaning, we introduce attention mechanism (i.e., node-level attention) to highlight those nodes that are important to the meaning of the hyperedge and then aggregate them to compute the hyperedge representation $\mathbf{f}_j^l$. Formally:
\begin{equation}
    \mathbf{f}_j^l = \sigma\bigg( \sum_{v_k \in e_j} \alpha_{jk} \mathbf{W}_{1} \mathbf{h}_{k}^{l-1}\bigg),
\end{equation}
where $\sigma$ is the nonlinearity such as ReLU and $\mathbf{W}_1$ is a trainable weight matrix. $\alpha_{jk}$ denotes the attention coefficient of node $v_k$ in the hyperedge $e_j$, which can be computed by:
\begin{equation}
\begin{aligned}
    \alpha_{jk} &= \frac{\exp(\mathbf{a}^{\mathrm{T}}_1 \mathbf{u}_{k})}
     {\sum_{v_p \in e_j}\exp(\mathbf{a}^{\mathrm{T}}_1 \mathbf{u}_{p})},\\
     \mathbf{u}_{k} &= \text{LeakyReLU}(\mathbf{W}_{1} \mathbf{h}_{k}^{l-1}),
\end{aligned}
\label{equ:att_node2edge}
\end{equation}
where $\mathbf{a}^{\mathrm{T}}_1$ is a weight vector (a.k.a, context vector).

\smallskip
\paragraph{Edge-level Attention.} With all the hyperedges representations $\{\mathbf{f}_j^{l} \arrowvert \forall e_j \in \mathcal{E}_i\}$, we again apply an edge-level attention mechanism to highlight the informative hyperedges for learning the next-layer representation of node $v_i$. This process can be formally expressed as:
\begin{equation}
    \mathbf{h}_{i}^{l} = \sigma\bigg( \sum_{e_j \in \mathcal{E}_i} \beta_{ij} \mathbf{W}_2 \mathbf{f}_j^l \bigg), 
\end{equation}
where $\mathbf{h}_{i}^{l}$ is the output representation of node $v_i$ and $\mathbf{W}_2$ is a weight matrix. $\beta_{ij}$ denotes the attention coefficient of hyperedge $e_j$ on node $v_i$, which can be computed by: 
\begin{equation}
\begin{aligned}
    \beta_{ij} &= \frac{\exp(\mathbf{a}^{\mathrm{T}}_2 \mathbf{v}_{j})}
     {\sum_{e_p \in \mathcal{E}_i}\exp(\mathbf{a}^{\mathrm{T}}_2 \mathbf{v}_{p})},\\
    \mathbf{v}_{j} &= \text{LeakyReLU}( [\mathbf{W}_{2}\mathbf{f}_j^l || \mathbf{W}_{1}\mathbf{h}_i^{l-1}]),
\end{aligned}
\label{equ:att_edge2node}
\end{equation}
where $\mathbf{a}^{\mathrm{T}}_2$ is another weight (context) vector for measuring the importance of the hyperedges and $||$ is the concatenation operation.

The proposed dual attention mechanism enables a HyperGAT layer not only to capture the high-order word interactions, but also to highlight the key information at different granularities during the node representation learning process.

% importance of different neighboring nodes (reducing the weights of noisy information) and the importance of different node (information) types to a specific node\JL{}.

\subsection{Inductive Text Classification}
For each document, after going through $L$ HyperGAT layers, we are able to compute all the node representations on the constructed text hypergraph. Then we apply the \textit{mean-pooling} operation on the learned node representations $\mathbf{H}^L$ to obtain the document representation $\mathbf{z}$, and feed it to a softmax layer for text classification. Formally:
\begin{equation}
    \hat{\mathbf{y}} = \text{softmax}\big(\mathbf{W}_c \mathbf{z}  + \mathbf{b}_c \big),
\end{equation}
where $\mathbf{W}_c$ is a parameter matrix mapping the document representation into an output space and $\mathbf{b}_c$ is the bias.
$\hat{\mathbf{y}}$ denotes the predicted label scores. Specifically, the loss function of text classification is defined as the cross-entropy loss:
%\mathcal{L} = - y_d \sum_d \text{log}(\hat{y}_d),
\begin{equation}
     \mathcal{L} = - \sum_d \text{log}(\hat{\mathbf{y}}^d_{j}),
\end{equation}
where $j$ is the ground truth label of document $d$. Thus HyperGAT can be learned by minimizing the above loss function over all the labeled documents.

%$y_d$ is the label of document $d$. 

Note that HyperGAT eliminates the mandatory access of test documents during training, making the model different from existing GNN-based methods. For unseen documents, we can directly feed their corresponding text hypergraphs to the previously learned model and compute their labels. Hence, we can handle the newly added data in an inductive way instead of retraining the model.

% Note that after the training phase, .
% To compute the anomaly scores of unseen nodes, we can retain the parameters of previously learned model and directly feed the new (sub)network
% \begin{equation}
%     \mathbf{Z}_l = FFN(\sigma(\sum_{w \in \mathbf{T}_l}\theta_{lw} \mathbf{W}\mathbf{h}_{w}^{L}))
% \end{equation}

%  To calculate the attention scores:
% \begin{equation}
%      \theta_{lw} = \frac{\exp(\sigma(\mathbf{a}^T\mathbf{W} \mathbf{h}_{w}^{L}))}
%      {\sum_{v \in \mathbf{T}_l}\exp(\sigma(\mathbf{a}^T\mathbf{W} \mathbf{h}_{v}^{L}))}
% \end{equation}

\section{Experiments}

\begin{table}[t!]
\centering
% \vspace{-0.1in}
\scalebox{0.825}{
\begin{tabular}{@{}c|ccccc@{}}
\toprule
\textbf{Dataset} & \textbf{20NG} & \textbf{R8} & \textbf{R52} & \textbf{Ohsumed}  & \textbf{MR}\\ \midrule
\textbf{\# Doc} & 18,846  & 7,674 & 9,100 & 7,400 & 10,662\\
\textbf{\# Train} & 11,314  &  5,485 &  6,532 &  3,357 &  7,108\\
\textbf{\# Test} & 7,532 & 2,189 & 2,568 &  4,043 & 3,554  \\
\textbf{\# Word} & 42,757 & 7,688 &  8,892 &   14,157 &  18,764  \\
\textbf{Avg Len} & 221.26  & 65.72 & 69.82 & 135.82 & 20.39 \\
\textbf{\# Class} & 20  & 8 & 52 & 23 & 2 \\
\bottomrule
\end{tabular}}
\caption{Summary statistics of the evaluation 
datasets.}
\label{tab:dataset}
\end{table}

\begin{table*}[t!]
% \vspace{-0.1in}
\centering
\scalebox{0.825}{
\begin{tabular}{@{}c|ccccc@{}}
\toprule
\textbf{Model} & \textbf{20NG} & \textbf{R8} & \textbf{R52} & \textbf{Ohsumed}  & \textbf{MR}\\ \midrule
% TF-IDF + LR & 0.8319 $\pm$ 0.0000 & 0.9374 $\pm$ 0.0000 & 0.8695 $\pm$ 0.0000 & 0.5466 $\pm$ 0.0000 & 0.7459 $\pm$ 0.0000\\
CNN-rand & 0.7693 $\pm$ 0.0061 & 0.9402 $\pm$ 0.0057 & 0.8537 $\pm$ 0.0047 & 0.4387 $\pm$ 0.0100 & 0.7498 $\pm$ 0.0070\\
CNN-non-static & 0.8215 $\pm$ 0.0052 & 0.9571 $\pm$ 0.0052 & 0.8759 $\pm$ 0.0048 & 0.5833 $\pm$ 0.0106 & 0.7775 $\pm$ 0.0072\\
LSTM & 0.6571 $\pm$ 0.0152 & 0.9368 $\pm$ 0.0082 & 0.8554 $\pm$ 0.0113 & 0.4114 $\pm$ 0.0117 & 0.7506 $\pm$ 0.0044\\
LSTM (pretrain) & 0.7543 $\pm$ 0.0172 & 0.9609 $\pm$ 0.0019 & 0.9048 $\pm$ 0.0086 & 0.5110 $\pm$ 0.0150 & 0.7733 $\pm$ 0.0089\\
Bi-LSTM & 0.7318 $\pm$ 0.0185 & 0.9631 $\pm$ 0.0033 & 0.9054 $\pm$ 0.0091 & 0.4927 $\pm$ 0.0107 & 0.7768 $\pm$ 0.0086\\
% PV-DBOW & 0.7436 $\pm$ 0.0018 & 0.8587 $\pm$ 0.0010 & 0.7829 $\pm$ 0.0011 & 0.4665 $\pm$ 0.0019 \\
% PV-DM & 0.5114 $\pm$ 0.0022 & 0.5207 $\pm$ 0.0004 & 0.4492 $\pm$ 0.0005 & 0.2950 $\pm$ 0.0007 \\
% PTE & 0.7476 $\pm$ 0.0029 & 0.9669 $\pm$ 0.0013 & 0.9071 $\pm$ 0.0014 & 0.5358 $\pm$ 0.0029 & 0.7023 $\pm$ 0.0036\\
fastText & 0.7938 $\pm$ 0.0030 & 0.9613 $\pm$ 0.0021 & 0.9281 $\pm$ 0.0009 & 0.5770 $\pm$ 0.0049 & 0.7514 $\pm$ 0.0020\\
fastText (bigrams) & 0.7967 $\pm$ 0.0029 & 0.9474 $\pm$ 0.0011 & 0.9099 $\pm$ 0.0005 & 0.5569 $\pm$ 0.0039 & 0.7624 $\pm$ 0.0012\\
SWEM & 0.8516 $\pm$ 0.0029 & 0.9532 $\pm$ 0.0026 & 0.9294 $\pm$ 0.0024 & 0.6312 $\pm$ 0.0055 & 0.7665 $\pm$ 0.0063\\
LEAM & 0.8191 $\pm$ 0.0024 & 0.9331 $\pm$ 0.0024 & 0.9184 $\pm$ 0.0023 & 0.5858 $\pm$ 0.0079 & 0.7695 $\pm$ 0.0045\\
\midrule
Graph-CNN & 0.8142 $\pm$ 0.0032 & 0.9699 $\pm$ 0.0012 & 0.9275 $\pm$ 0.0022 & 0.6386 $\pm$ 0.0053 & 0.7722 $\pm$ 0.0027\\
% Graph-CNN-S & - & 0.9680 $\pm$ 0.0020 & 0.9274 $\pm$ 0.0024 & 0.6282 $\pm$ 0.0037 & 0.7699 $\pm$ 0.0014\\
% Graph-CNN-F & - & 0.9689 $\pm$ 0.0006 & 0.9320 $\pm$ 0.0004 & 0.6304 $\pm$ 0.0077 & 0.7674 $\pm$ 0.0021\\
TextGCN (transductive) & 0.8643 $\pm$ 0.0009 & 0.9707 $\pm$ 0.0010 & 0.9356 $\pm$ 0.0018 & 0.6836 $\pm$ 0.0056 & 0.7674 $\pm$ 0.0020\\

TextGCN (inductive) & 0.8331 $\pm$ 0.0026 & 0.9578 $\pm$ 0.0029 & 0.8820 $\pm$ 0.0072 & 0.5770 $\pm$ 0.0035 & 0.7480 $\pm$ 0.0025\\

Text-level GNN  & 0.8416 $\pm$ 0.0025 & 0.9789 $\pm$ 0.0020 & 0.9460 $\pm$ 0.0030 & 0.6940 $\pm$ 0.0060 & 0.7547 $\pm$ 0.0006 \\\midrule

% TextGAT (ours) & 0.8555 $\pm$ 0.0002 & 0.9688 $\pm$ 0.0029 & 0.9299 $\pm$ 0.0036 & 0.6527 $\pm$ 0.0068 & 0.7673 $\pm$ 0.0040\\
HyperGAT (ours) & 0.8662 $\pm$ 0.0016 & 0.9797 $\pm$ 0.0023 & 0.9498 $\pm$ 0.0027 & 0.6990 $\pm$ 0.0034 & 0.7832 $\pm$ 0.0027\\
\bottomrule
\end{tabular}}

\caption{Test accuracy on document classification with different models. Each model we ran 10 times and report the mean $\pm$ standard deviation. HyperGAT significantly outperforms all the baselines based on t-tests ($p<0.05$).}
\label{tab:result}
\end{table*}

% $\ast$ indicates that the improvement of the best result is statistically significant compared with the next-best result with $p<0.01$.

\subsection{Experimental Setting}
\smallskip
\paragraph{Evaluation Datasets.}
To conduct a fair and comprehensive evaluation, we adopt five benchmark datasets from different domains in our experiments: 20-Newsgroups (20NG), Reuters (R8 and R52), Ohsumed, and Movie Review (MR). Those datasets have been widely used for 
evaluating graph-based text classification performance~\cite{yao2019graph,huang2019text,liu2020tensor}. Specifically, the 20-Newsgroups dataset and two Reuters datasets are used for news classification. The Ohsumed dataset is medical literature. The Movie Review dataset is collected for binary sentiment classification. A summary statistics of the benchmark datasets is presented in table \ref{tab:dataset} and more detailed descriptions can be found in~\cite{yao2019graph}. For quantitative evaluation, we follow the same train/test splits and data preprocessing procedure in~\cite{yao2019graph} in our experiments. In each run, we randomly sample 90\% of the training samples to train the model and 
use the left 10\% data for validation. More details can be found in Appendix A.1.

\smallskip
\paragraph{Compared Methods.}
In our experiments, the baselines compared with our model HyperGAT can be generally categorized into three classes: (i) \textit{word embedding-based} methods that classify documents based on pre-trained word embeddings, including fastText~\cite{joulin2016bag}, and more advanced methods SWEM~\cite{Shen2018Baseline} and LEAM~\cite{wang2018joint}; (ii) \textit{sequence-based} methods which capture text features from local consecutive word sequences, including CNNs~\cite{kim2014convolutional}, LSTMs~\cite{liu2016recurrent}, and Bi-LSTM~\cite{huang2015bidirectional}; (iii) \textit{graph-based} methods that aim to capture interactions between words, including Graph-CNN~\cite{defferrard2016convolutional}, two versions of TextGCN~\cite{yao2019graph} and Text-level GNN~\cite{huang2019text}. Note that TextGCN (transductive) is the model proposed in the original paper and TextGCN (inductive) is the inductive version implemented by the same authors. Text-level GNN is a state-of-the-art baseline which performs text representation learning on document-level graphs. More details of baselines can be found in~\cite{yao2019graph}.

\smallskip
\paragraph{Implementation Details.} HyperGAT is implemented by PyTorch and optimized with the Adam optimizer. We train and test the model on a 12 GB Titan Xp GPU. Specifically, our HyperGAT model consists of two layers with 300 and 100 embedding dimensions, respectively. We use one-hot vectors as the node attributes and the batch size is set to 8 for all the datasets. The optimal values of hyperparameters are selected when the model achieves the highest accuracy for the validation samples. The optimized learning rate $\alpha$ is set to 0.0005 for MR and 0.001 for the other datasets. L2 regularization is $10^{-6}$ and dropout rate is 0.3 for the best performance. For learning HyperGAT, we train the model for 100 epochs with early-stopping strategy. To construct the semantic hyperedges, we train an LDA model for each dataset using the training documents and select the Top-10 words from each topic. The topic number is set to the same number of classes. For baseline models, we either show the results reported in previous research~\cite{yao2019graph} or run the codes provided by the authors using the parameters described in the original papers. \textit{More details can be found in the Appendix A.2.} Our data and source code is available at \url{https://github.com/kaize0409/HyperGAT}.

\subsection{Experimental Results}

\smallskip
\paragraph{Classification Performance.}
We first conduct comprehensive experiments to evaluate model performance on text classification and present the results in Table \ref{tab:result}. Overall, our model HyperGAT outperforms all the baselines on the five evaluation datasets, which demonstrates its superior capability in text classification. In addition, we can make the following in-depth observations and analysis:
\begin{itemize}[leftmargin=*,itemsep=1pt,topsep=1.5pt]
\item \textit{Graph-based} methods, especially GNN-based models are able to achieve superior performance over the other two categories of baselines on the first four datasets. This observation indicates that text classification performance can be directly improved by capturing long-distance word interactions. While for the MR dataset, \textit{sequence-based} methods (CNNs and LSTMs) show stronger classification capability than most of the \textit{graph-based} baselines. One potential reason is that sequential context information plays a critical role in sentiment classification, which cannot be explicitly captured by the majority of existing \textit{graph-based} methods.

\item Not surprisingly, without the additional knowledge on test documents, the performance of TextGCN (inductive) largely falls behind its original transductive version. Though Text-level GNN is able to achieve performance improvements by adding trainable edge weights between word, its performance is still limited by the information loss of using pairwise simple graph. In particular, our model HyperGAT achieves considerable improvements over other GNN-based models, demonstrating the importance of high-order context information for learning word representations.

% When pre-trained GloVe word embeddings are provided, CNN performs much better, especially on Ohsumed and 20NG. CNN also achieves the best results on short text dataset MR with pre-trained word embeddings, which shows it can.

% \item  Another potential reason is that the edges in MR text graph are fewer than other text graphs, which limits the message passing among the nodes. There are only few document-word edges because the documents are very short. 

% Because the short documents in MR lead to a lowdensity graph in TextGCN, it restrains the label message passing among document nodes, whereas our individual graphs (documents) do not rely on such label message passing mechanism

\end{itemize}

\begin{table}
\centering
\vspace{0.2cm}
\scalebox{0.825}{
\begin{tabular}{@{}c|ccc@{}}
\toprule
\textbf{Model} & \textbf{TextGCN (transductive)} &  \textbf{HyperGAT} \\ \midrule
\textbf{20NG} & 1,4479.36MB  &  180.33MB  \\
\textbf{R8} & 931.58MB  &  41.75MB  \\
\textbf{R52} & 1289.48MB  &  46.85MB  \\
\textbf{Ohsumed} & 1822.71MB  &  63.17MB  \\
\textbf{MR} & 3338.24MB &80.99MB  \\
\bottomrule
\end{tabular}}
\caption{GPU memory consumption of different methods. The batch size for HyperGAT is set to 8. }
\label{tab:memory}
\end{table}

\smallskip
\paragraph{Computational Efficiency.}
Table \ref{tab:memory} presents the computational cost comparison between the most representative transductive baseline TextGCN and our approach. Form the reported results, we can clearly find that HyperGAT has a significant computational advantage in terms of memory consumption. The main reason is that HyperGAT conducts text representation learning at the document-level and it only needs to store a batch of small text hypergraphs during training. On the contrary, TextGCN requires constructing a large document-word graph using both training and test documents, which inevitably consumes a great amount of memory. Another computational advantage of our model is that HyperGAT is an inductive model that can generalize to unseen documents. Thus we do not have to retrain the whole model for newly added documents like transductive methods.

\begin{figure}[t]
    \graphicspath{{figures/}}
    \centering
    \subfigure[\textbf{20NG}] 
    {
    \includegraphics[width=0.48\columnwidth]{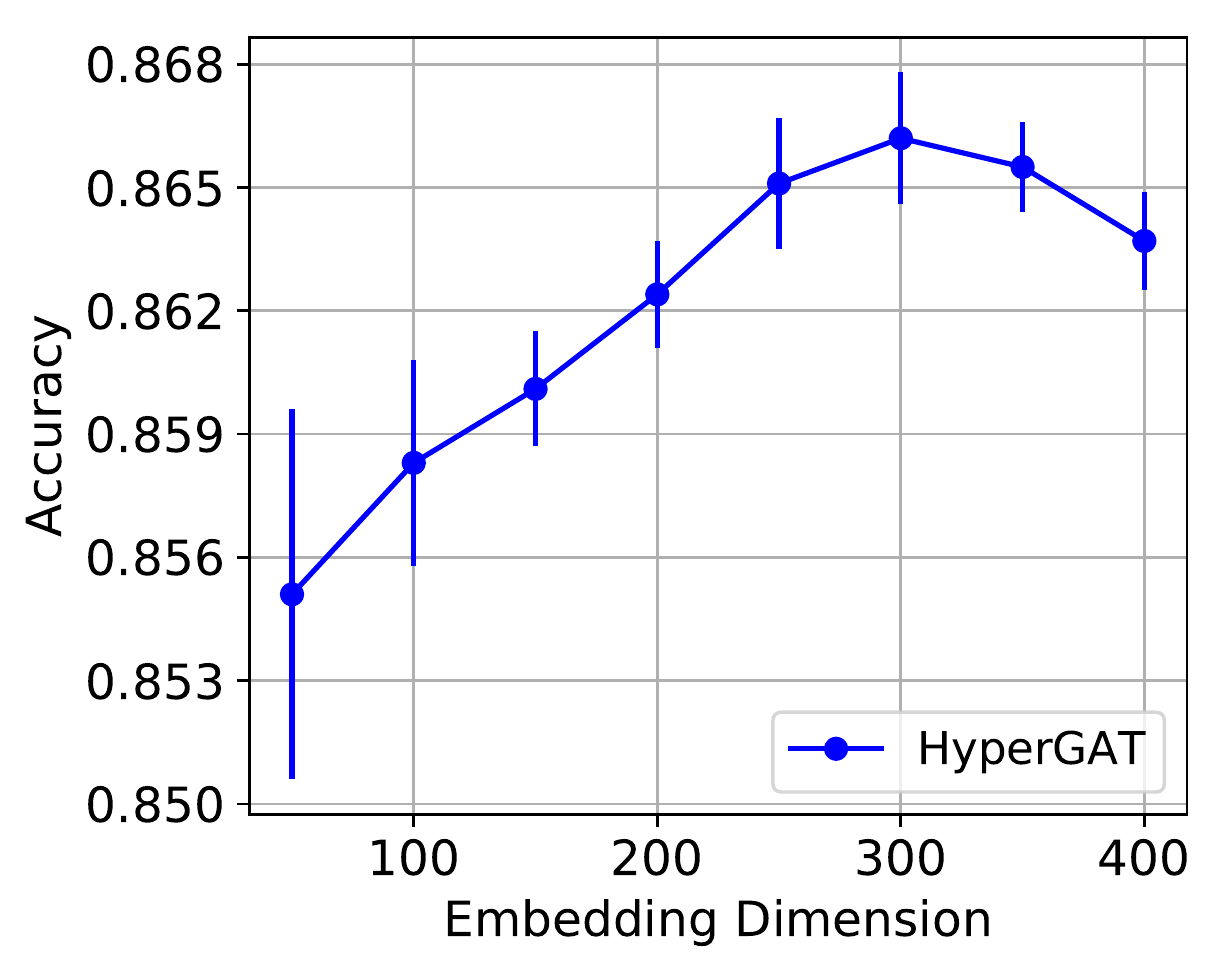}
    }
    \hspace{-0.3cm}
    \subfigure[\textbf{Ohsumed}]
    {
    \includegraphics[width=0.48\columnwidth]{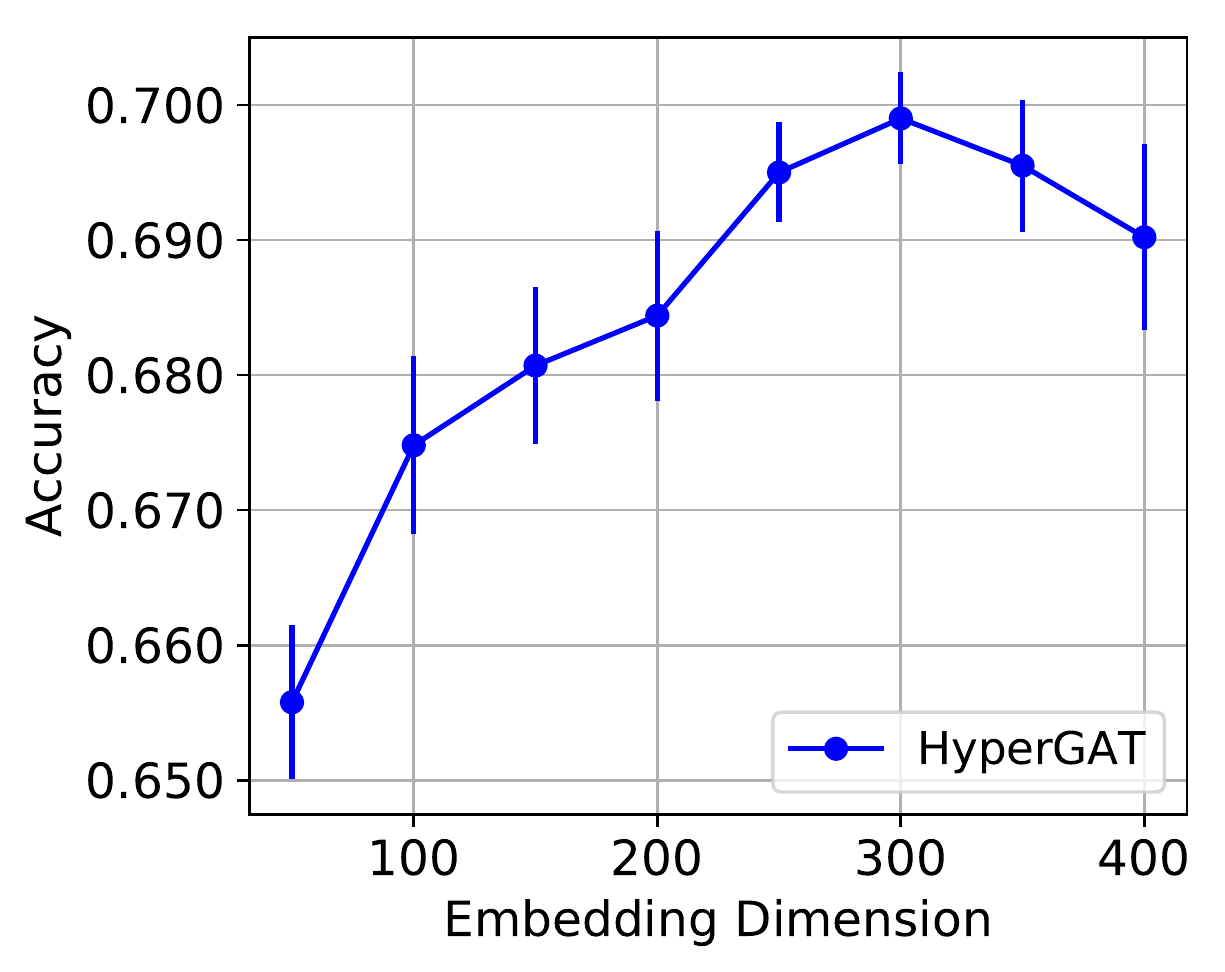}
    }
    \caption{Test accuracy by varying the embedding size of the first HyperGAT layer.}
    \label{fig:embedding}
\end{figure}

\begin{figure}[t!]
    \graphicspath{{figures/}}
    \centering
    \subfigure[\textbf{Ohsumed}] 
    {
    \includegraphics[width=0.48\columnwidth]{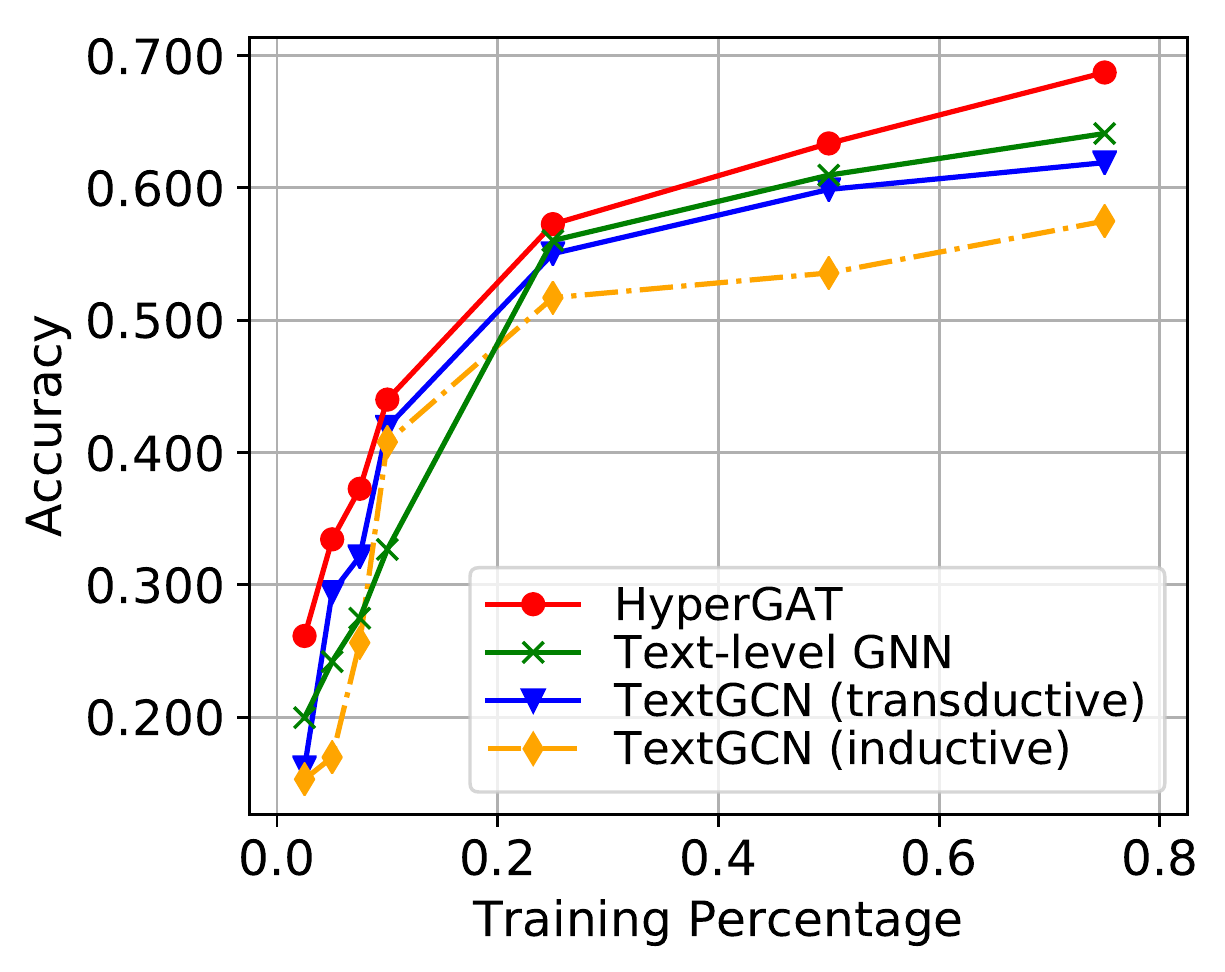}
    }
    \hspace{-0.3cm}
    \subfigure[\textbf{MR}]
    {
    \includegraphics[width=0.48\columnwidth]{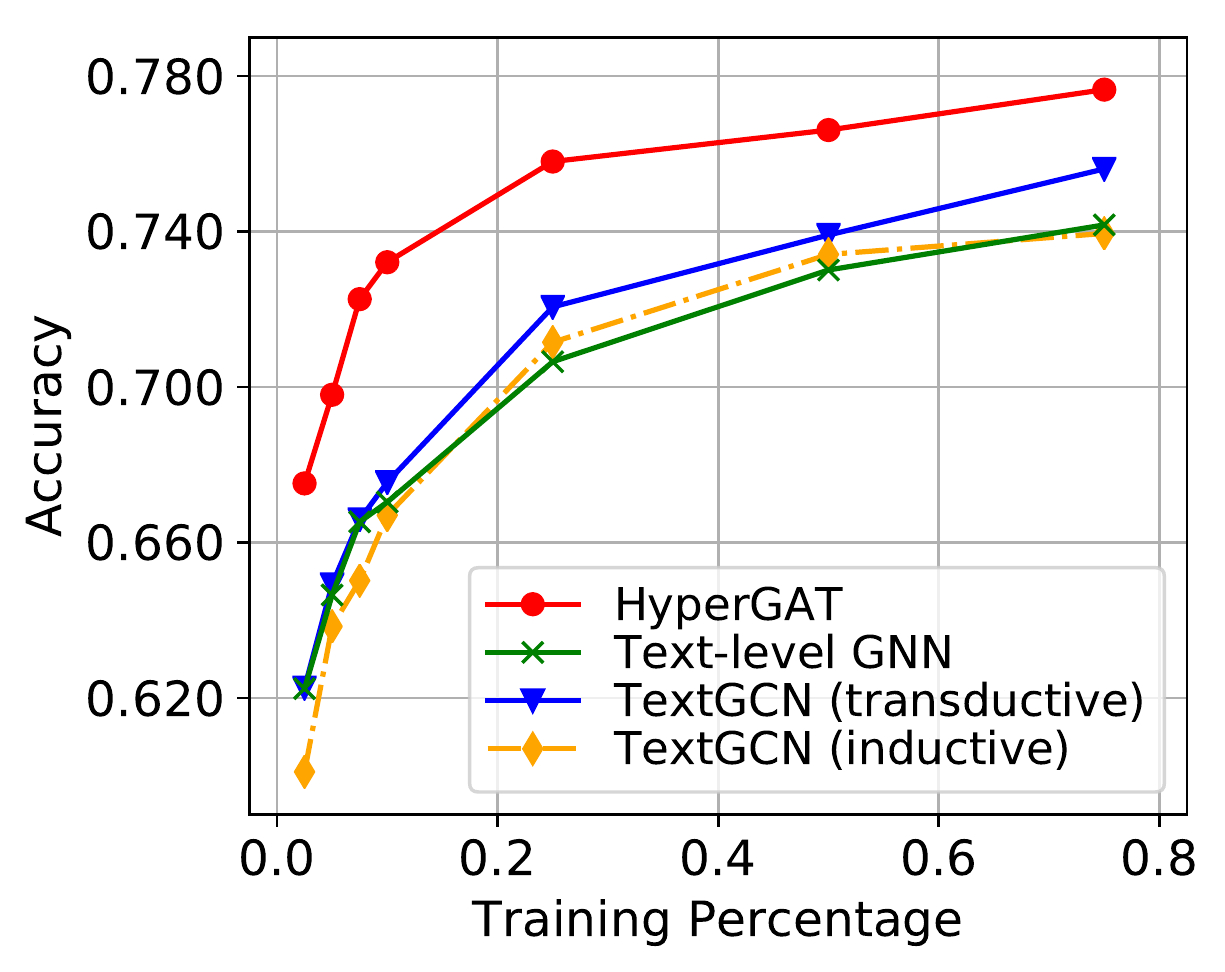}
    }
    \caption{Test accuracy by varying the proportions of training data (2.5\%, 5\%, 7.5\%, 10\%, 25\%, 50\%, 75\%).}
    \label{fig:training}
\end{figure}

\smallskip
\paragraph{Model Sensitivity.} 
The model performance on 20NG and
Ohsumed with different first-layer embedding dimensions is reported in Figure \ref{fig:embedding}, and we omit the results on other datasets since similar results can be observed. Notably, the best performance of HyperGAT is achieved when the first-layer embedding size is set to 300. It indicates that small embedding size may render the model less expressive, while the model may encounter overfitting if the embedding size is too large. In the meantime, to evaluate the effect of the size of labeled training data, we compare several best performing models with different proportions of the
training data and report the results on Ohsumed and MR in Figure \ref{fig:training}. In general, with the growth of labeled training data, all the evaluated methods can achieve performance improvements. More remarkably, HyperGAT can significantly outperform other baselines with limited labeled data, showing its effectiveness in real-world scenarios.

\begin{table*}[t]
\centering
\scalebox{0.825}{
\begin{tabular}{@{}c|ccccc@{}}
\toprule
\textbf{Model} & \textbf{20NG} & \textbf{R8} & \textbf{R52} & \textbf{Ohsumed} & \textbf{MR}\\ \midrule
% Text-GAT & 0.8555 $\pm$ 0.0002 & 0.9688 $\pm$ 0.0029 & 0.9299 $\pm$ 0.0036 & 0.6527 $\pm$ 0.0068 & 0.7673 $\pm$ 0.0040\\
w/o attention & 0.8645 $\pm$ 0.0006 & 0.9705 $\pm$ 0.0015 & 0.9321 $\pm$ 0.0023 & 0.6611 $\pm$ 0.0042 & 0.7699 $\pm$ 0.0044 \\
% Average Aggregation & 0.8605 $\pm$ 0.0021 & 0.9708 $\pm$ 0.0023 & 0.9399 $\pm$ 0.0030 & 0.6810 $\pm$ 0.0056 & 0.7795 $\pm$ 0.0025\\
w/o sequential & 0.6813 $\pm$ 0.0024 & 0.9448 $\pm$ 0.0053 &  0.9051 $\pm$ 0.0023 & 0.5664 $\pm$ 0.0047 & 0.7766 $\pm$ 0.0009\\
w/o semantic & 0.8602 $\pm$ 0.0031 & 0.9714 $\pm$ 0.0026 & 0.9415 $\pm$ 0.0032 & 0.6848 $\pm$ 0.0045 & 0.7811 $\pm$ 0.0028\\
HyperGAT (1 layer) & 0.8610 $\pm$ 0.0014 & 0.9735 $\pm$ 0.0012 & 0.9472 $\pm$ 0.0023 & 0.6913 $\pm$ 0.0023 & 0.7788 $\pm$ 0.0016 \\
HyperGAT & 0.8662 $\pm$ 0.0016 & 0.9797 $\pm$ 0.0023 & 0.9498 $\pm$ 0.0027 & 0.6990 $\pm$ 0.0034 & 0.7832 $\pm$ 0.0027\\
\bottomrule
\end{tabular}}
\caption{Text classification comparison results \textit{w.r.t.} test accuracy (mean $\pm$ standard deviation). HyperGAT significantly outperforms all its variants on each dataset based on t-tests ($p<0.05$).}
\label{tab:ablation}
\end{table*}

\subsection{Ablation Analysis}
To investigate the contribution of each module in HyperGAT, we conduct an ablation analysis and report the results in Table \ref{tab:ablation}. Specifically, \textit{w/o attention} is a variant of HyperGAT that replaces the dual attention with convolution. \textit{w/o sequential} and \textit{w/o semantic} are another two variants by excluding sequential, semantic hyperedges, respectively. 
From the reported results we can learn that HyperGAT can achieve better performance by stacking more layers. This observation can verify the usefulness of long-distance word interactions for text representation learning. Moreover, the performance gap between \textit{w/o attention} and HyperGAT shows the effectiveness of the dual attention mechanism for learning more expressive word representations. By comparing the results of \textit{w/o sequential} and \textit{w/o semantic}, we can learn that the context information encoded by the sequential hyperedges is more important, but adding semantic hyperedges can enhance the model expressiveness. It also indicates that heterogeneous high-order context information can complement each other and we could investigate more meaningful hyperedges to further improve the performance of our approach.

\begin{figure}[b]
    \graphicspath{{figures/}}
    \centering
    \vspace{-0.5cm}
    \subfigure[\textbf{Text-level GNN}]
    {
    \includegraphics[width=0.485\columnwidth]{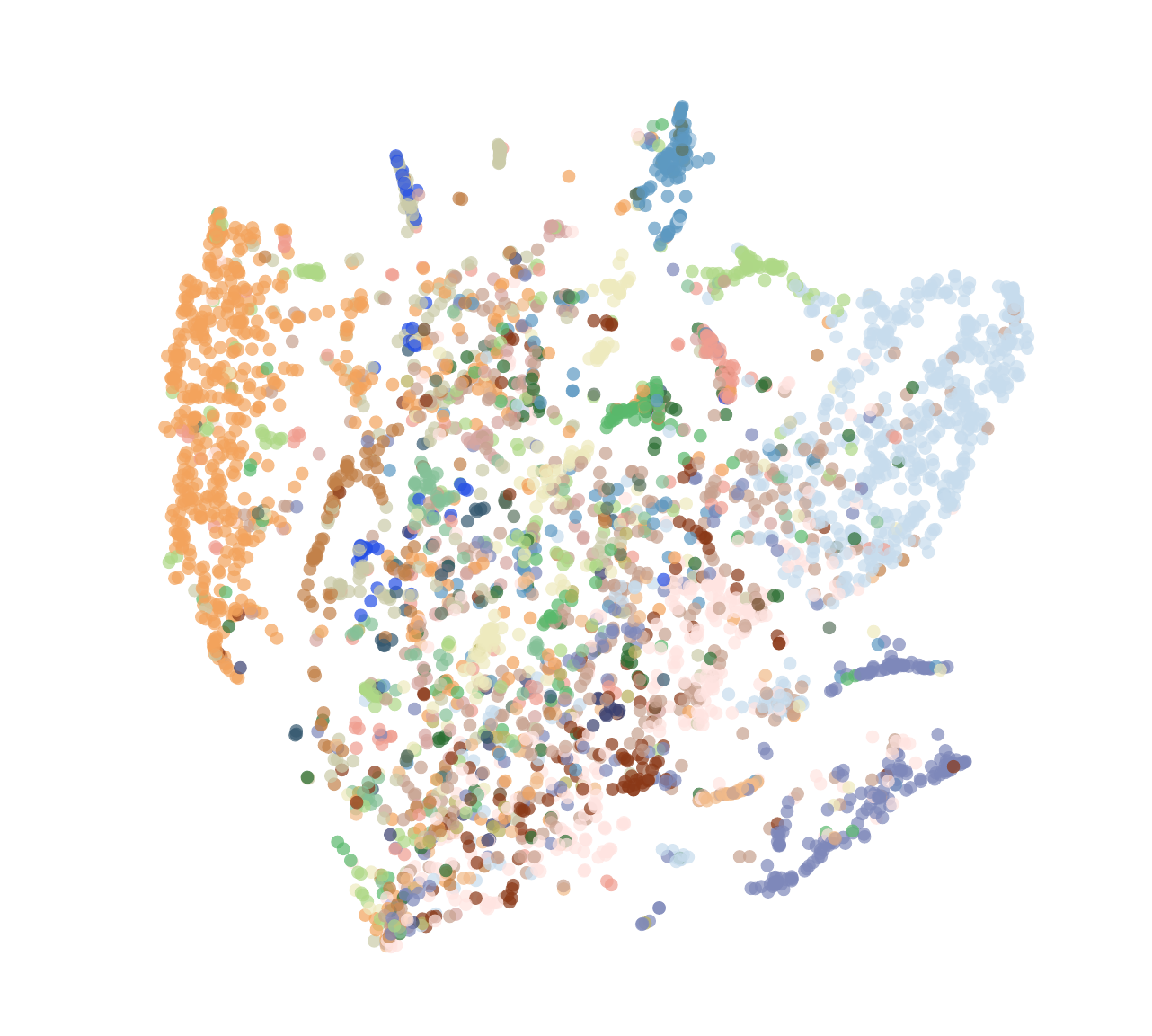}
    }
    \hspace{-0.5cm}
    \subfigure[\textbf{HyperGAT}] 
    {
    \includegraphics[width=0.485\columnwidth]{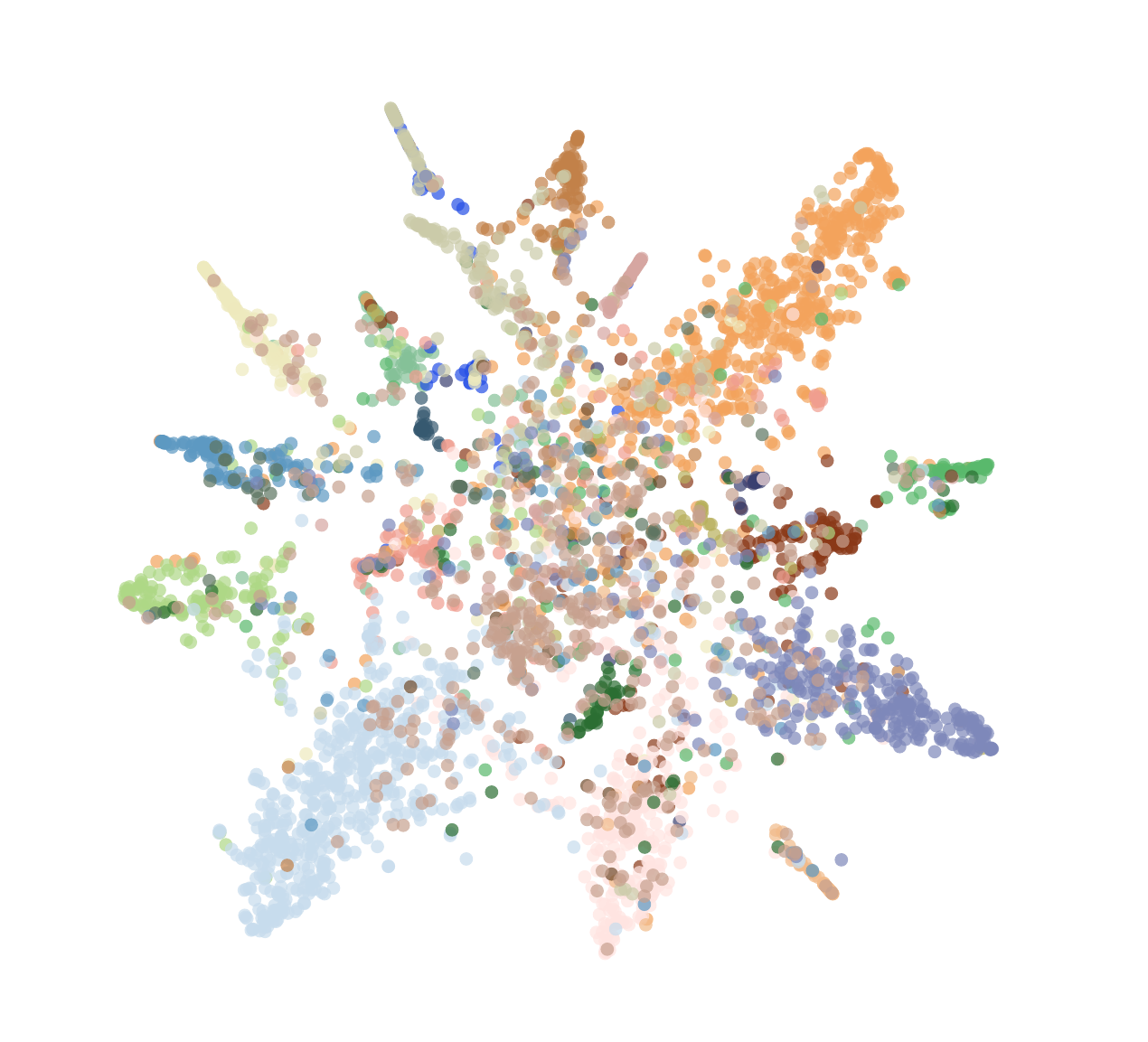}
    }
    
    % \quad

    % \hspace{-0.5cm}
    % \subfigure[\textbf{TextGCN (transductive)}]
    % {
    % \includegraphics[width=0.485\columnwidth]{emnlp2020-templates/figures/viz_textgcn_trans.pdf}
    % }
    \caption{The t-SNE visualization of Text-level GNN and HyperGAT for test documents in Ohsumed.}%

    \label{fig:tsne}
\end{figure} 

\subsection{Case Study}
\smallskip
\paragraph{Embedding Visualization.}
In order to show the superior embedding quality of HyperGAT over other methods, we use t-SNE~\cite{maaten2008visualizing} to visualize the learned representations of documents for comparison. Specifically, Figure \ref{fig:tsne} shows the visualization results of the best performing baseline Text-level GNN and HyperGAT on the test documents of Ohsumed. Note that the node’s color corresponds to its label, which is used to verify the model's expressive power on 23 document classes. From the embedding visualization, we are able to observe that HyperGAT can learn more expressive document representations over the state-of-the-art method Text-level GNN.

% Due to the space
% limitation, here we only post the results on Ohsumed dataset. 

\begin{figure}[t]
\includegraphics[width=0.47\textwidth]{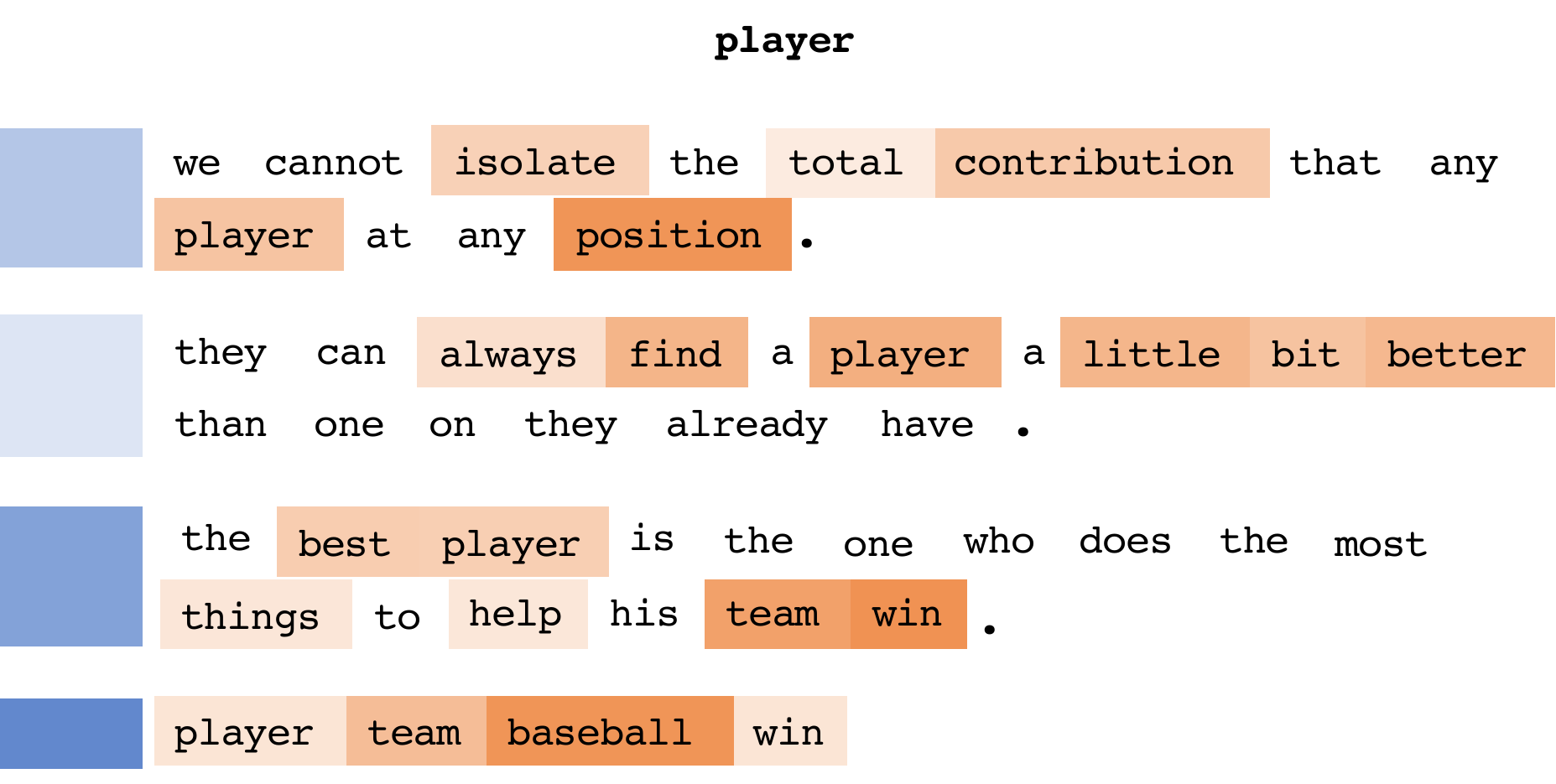}
\caption{Visualization of the dual attention mechanism in HyperGAT. Figure best viewed in color.}
\label{fig:case1}
\end{figure}

\smallskip
\paragraph{Attention Visualization.}
% We take a short text from AGNews as an example (which is classified to the
% class of sports correctly) to illustrate the duallevel attention of HGAT
To better illustrate the learning process of the proposed dual attention mechanism, we take a text document from 20NG (labeled as \textit{sport.baseball} correctly) and visualize the attention weights computed for the word \texttt{player}. As shown in Figure \ref{fig:case1}, \texttt{player} is connected to four hyperedges within the constructed document-level hypergraph. The first three lines ended with periods represent sequential hyperedges, while the last one without a period is a semantic hyperedge. Note that we use orange to denote the node-level attention weight and blue to denote the edge-level attention weight. Darker color represents larger attention weight.

On the one hand, node-level attention is able to 
select those nodes (words) carrying informative context on the same hyperedge. For example, \texttt{win} and \texttt{team} in the third hyperedge gain larger attention weights since they are more expressive compared to other words in the same sentence. On the other hand, edge-level attention can also assign fine-grained weights to highlight meaningful hyperedges. As we can see, the last hyperedge that connects \texttt{player} with \texttt{baseball} and \texttt{win} receives higher attention weight since it can better characterize the meaning of \texttt{player} in the document. To summarize, this case study shows that our proposed dual attention can capture key information at different granularities for learning expressive text representations.

\section{Conclusion}
In this study, we propose a new graph-based method for solving the problem of inductive text classification. Apart from the existing efforts, we propose to model text documents with document-level hypergraphs and further develop a new family of GNN model named HyperGAT for learning discriminative text representations. Specifically, our method is able to acquire more expressive power with less computational consumption for text representation learning. By conducting extensive experiments, the results demonstrate the superiority of the proposed model over the state-of-the-art methods.

\section*{Acknowledgements}
This material is in part supported by the
National Science Foundation (NSF) grant 1614576.

\balance
\bibliography{emnlp2020}
\bibliographystyle{acl_natbib}

\appendix
\section{Appendix}

%\begin{table}[!h]
%\centering
%\caption{Links for accessing the evaluation datasets}
%\resizebox{0.48\textwidth}{!}{%
%\begin{tabular}{ll}
%\toprule
 %\textbf{Datasets} & \multicolumn{1}{c}{\textbf{Links}} \\ \hline
%20ng & \url{http://qwone.com/~jason/20Newsgroups/} \\ \hline
%R52 &  \url{http://www.daviddlewis.com/resources/testcollections/}\\ \hline
%R8 & \url{http://www.daviddlewis.com/resources/testcollections/} \\ \hline
%Ohsumed & \url{http://disi.unitn.eu/moschitti/corpora.htm} \\ \hline
%MR & \url{https://www.cs.cornell.edu/people/pabo/movie-review-data/} \\ 
%\bottomrule
%\end{tabular}
%}
%\label{table:appendix_dataset}
%\end{table} 
%\begin{tabular}[c]{@{}l@{}}\end{tabular}

\subsection{Implementation Details}
\label{sec:reproducibility}
As the supplement to Section 4, in the following, we explain the implementation of HyperGAT.

 %and the compared methods in detail. The implementation of the proposed model can be found on \url{https://github.com/HyperGAT/HyperGAT}.

\smallskip
\noindent\textbf{LDA Model Training.} We use the implementation provided in scikit-learn to train the LDA model. We only use the documents in the training set to train the LDA model for each dataset. We select to use the Online Variational Bayes method for model learning. We set the random state is set to be 0 and the learning offset to be 50. As for the other parameters, we follow the default setting provided by scikit-learn. The topic number is set to be the same as the number of classes for each of the datasets. And we select the Top-10 keywords of each topic to construct the semantic hyperedges.

\smallskip
\noindent\textbf{Implementation of HyperGAT.} The proposed HyperGAT model is implemented in PyTorch and optimized with the Adam optimizer~\cite{kingma2014adam}. It is trained and tested on a 12 GB Titan Xp GPU. Specifically, the hypergraph attention network consists of two layers with 300 and 100 embedding dimensions, respectively. We use one-hot vectors as the node attributes. The batch size is set to be 8 for all the datasets. We grid search for the learning rate in \{0.0001, 0.0005, 0.001, 0.005, 0.01, 0.05, 0.1\}, L2 regularization in \{$10^{-6}$, $10^{-5}$, $10^{-4}$, $10^{-3}$, $10^{-2}$, $10^{-1}$\} and the dropout rate in \{0.1, 0.2, 0.3, 0.4, 0.5, 0.6, 0.7\}. The optimal values are selected when the model achieves the highest accuracy for the validation samples. The optimized learning rate $\alpha$ for MR is 0.0005 while that for the other datasets is 0.001. We select the L2 regularization to be $10^{-6}$ and the dropout rate to be 0.3 for the best performance. For each dataset, we train the model for 100 epochs or stop if the performance for the validation doesn't increase for 5 consecutive epochs as an early-stopping strategy. Under the optimized setup, the model can converge in 587s, 145s, 156s, 97s and 78s on average for 20NG, R8, R52, Ohsumed and MR, respectively.

\smallskip
\noindent\textbf{Validation Performance.} As supplement to the test results in Table \ref{tab:result}, we also report the corresponding validation performance of the proposed HyperGAT. The validation accuracy is $0.9355 \pm 0.0011$, $0.9755 \pm 0.0019$, $0.9375 \pm 0.0023$, $0.6964 \pm 0.0024$ and $0.7779 \pm 0.0015$ for 20NG, R8, R52, Ohsumed and MR, respectively.

\subsection{Space Complexity Analysis}
Theoretically, the main difference of memory usage between HyperGAT and other methods lies in the size of the adjacency matrix. Formally, let $N$ denote vocabulary size and $M$ denote document size. Take TextGCN as an example, the size of the adjacency matrix is $(N + M)^2$. As HyperGAT adopts document-level hypergraphs, for each hypergraph, the adjacency matrix size is $n \times m$, where $n$ is the number of words and $m$ is the number of hyperedges in a document. Based on mini-batch training, the memory consumption for each mini-batch is about $n \times m \times bsz$. Since $N$ and $M$ are way larger than $n$ and $m$, the memory consumption of HyperGAT can be largely reduced in practice.

\end{document}